\documentclass[11pt,a4paper]{article}
\usepackage[hyperref]{acl2019}
\usepackage{cc}

\def\biblio{\bibliographystyle{acl_natbib}\bibliography{cc}}

\aclfinalcopy

\title{Improving Long Distance Slot Carryover in Spoken Dialogue Systems}

\author{
  \bf Tongfei Chen$^*$ \quad
  Chetan Naik$^\dag$ \quad
  Hua He$^\dag$ \quad \\
  \bf Pushpendre Rastogi$^\dag$ \quad
  Lambert Mathias$^\dag$ \\
  $^*$ Johns Hopkins University \\
  $^\dag$ Amazon.com, Inc. \\
  {  \tt tongfei@jhu.edu, \string{chetnaik,huhe,prastogi,mathiasl\string}@amazon.com} \\
}

\date{}

\begin{document}
\def\biblio{}

\maketitle
\begin{abstract}
Tracking the state of the conversation is a central component in task-oriented spoken dialogue systems. One such approach for tracking the dialogue state is {\it slot carryover}, where a model makes a binary decision if a slot from the context is relevant to the current turn. Previous work on the slot carryover task used models that made independent decisions for each slot. A close analysis of the results show that this approach results in poor performance over longer context dialogues. In this paper, we propose to jointly model the slots. We propose two neural network architectures, one based on pointer networks that incorporate slot ordering information, and the other based on transformer networks that uses self attention mechanism to model the slot interdependencies. Our experiments on an internal dialogue benchmark dataset and on the public DSTC2 dataset demonstrate that our proposed models are able to resolve longer distance slot references and are able to achieve competitive performance.
\end{abstract}


\section{Introduction}
\label{sec:introduction}
\begin{figure}[ht]
  \centering
  \includegraphics[width=0.48\textwidth]{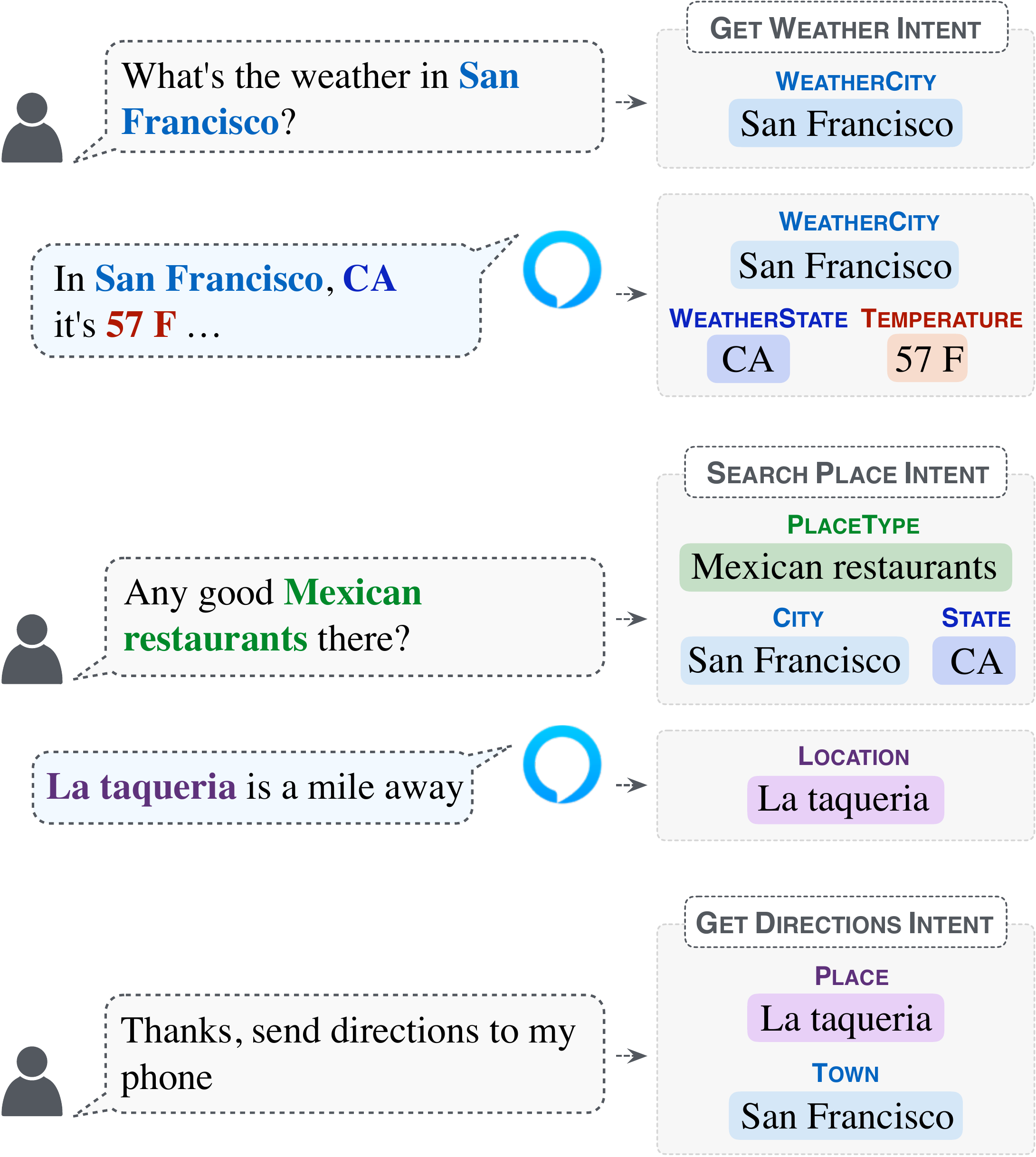}
  \vspace{-0.6cm}
  \caption{
    An example of a conversation session.
    Slots are listed on the right.
    Related slots often co-occur, such as (1) \slot{WeatherCity}{San Francisco} and \slot{WeatherState}{CA}, and should
    be carried over together due to their interdependencies (2) {\sc Place} slot is often seen to occur along with
    {\sc Town}.
  }
  \vspace{-0.3cm}
  \label{fig:dialogue}
\end{figure}

In task-oriented spoken dialogue systems, the user and the system are engaged in interactions that can span multiple turns.  A key challenge here is that the user can reference 
entities introduced in previous dialogue turns. For example, if a user request for \textit{what's the weather in arlington} is followed by \textit{how about tomorrow}, the dialogue system 
has to keep track of the entity \textit{arlington} being referenced.

In {\it slot-based} spoken dialogue systems, tracking the entities in context can be cast as {\it slot carryover} task -- only the relevant slots from the dialogue context are carried over to 
the current turn. Recent work by~\citet{naik2018contextual} describes a scalable multi-domain neural network
architecture to address the task in a diverse schema setting. However, this approach  treats every slot as independent. Consequently, as shown in our 
experiments, this results in lower performance when the contextual slot being referenced is associated with dialogue turns that are further away from the current turn. We posit 
that modeling slots jointly is essential for improving the accuracy over long distances, particularly when slots are correlated. We motivate this with an example conversation in 
\autoref{fig:dialogue}. In this example, the slots {\sc WeatherCity}/{\sc WeatherState}, need to be carried over {\it together} from dialogue history as they are correlated. However, the 
model in~\citet{naik2018contextual} has no information about this slot interdependence and may choose to carryover only one of the slots. In this work, we alleviate this issue by proposing two novel neural network architectures -- one based on pointer
networks \citep{vinyals2015pointer} and another based on self-attention with transformers \citep{vaswani2017attention} --  that can learn
to \emph{jointly} predict jointly whether a subset of related slots should be carried over from dialogue history. 

To validate our approach, we conduct thorough evaluations on both the publicly available  DSTC2 task \citep{henderson2014word}, as well as our internal 
dialogue dataset collected from a commercial digital assistant. In Section~\ref{sec:exp_results}, we show that our
proposed approach improve slot carryover accuracy over the baseline systems over longer dialogue contexts.  A detailed error analysis reveals that our proposed models are more likely to utilize ``anchor'' slots -- slots tagged in the current utterance -- to carry over long-distance slots from context. 

To summarize we make the following contributions in this work:
\begin{list}{\labelitemi}{\leftmargin=1em}
  \vspace{-0.1cm}
  \setlength{\itemsep}{-2pt}
  \item[1.] We improve upon the slot carryover model architecture in \citet{naik2018contextual} by introducing approaches for modeling slot interdependencies. We propose two neural 
network models based on pointer networks and transformer networks that can make joint predictions over slots.
  \item[2.] We provide a detailed analysis of the proposed models both on an internal benchmark and public dataset. We show that contextual encoding of slots and modeling slot 
interdependencies is essential for improving performance of slot carryover over longer dialogue contexts. Transformer architectures with self attention provide the best performance 
overall.
  \vspace{-0.1cm}
\end{list}

\biblio


\section{Problem Formulation}
\label{sec:problem_formulation}

A dialogue $H$ is formulated as a sequence of utterances, alternatively uttered by a user ($U$) and the system agent
($A$):
\begin{equation}
  H = \left(h_{d}^{\{U,A\}}, \cdots, h_2^{\rm U}, h_1^{\rm A}, h_0^{\rm U}\right) \ ,
\end{equation}
where each element $h$ is an utterance.
A subscript $d$ denotes the utterance distance which measures the offset from the most recent user utterance
($h^{\rm U}_{0}$). The $i$-th token of an utterance with distance $d$ is denoted as $h_d[i]$.

A slot $x = (d, k, l, r)$ in a dialogue is defined as a key-value pair that contains an entity information,
e.g. [{\it {\sc City:\textit{San Francisco}}}]. Each slot can be determined by the utterance distance $d$, slot key $k$,
and a span $[l:r]$ over the tokens of the utterance with slot value represented as $h_d[l:r]$.

Given a dialogue history $H$ and a set of candidate slots $X$, the context carryover task is addressed by deciding which
slots should be carried over. The previous work~\citep{naik2018contextual} addressed the task as a binary classification
problem and each slot $x \subseteq X$ is classified independently. In contrast, our proposed models can explicitly
capture slot interactions and make joint predictions of all slots. We show formulations of both model types below,

\begin{align}
  F_{\rm binary}(x, H) &\in (0, 1) \qquad \forall x \in X \\
  F_{\rm joint}(X, H) &\subseteq X
\end{align}
where $F_{\rm binary}(x, H)$ denotes a binary classification model~\citep{naik2018contextual}, $F_{\rm joint}(X, H)$
denotes our joint prediction models.

\biblio


\section{Models}
\label{sec:models}

\subsection{General architecture}
\label{ssec:general_architecture}
\begin{figure*}[ht]
  \centering
  \includegraphics[width=\textwidth]{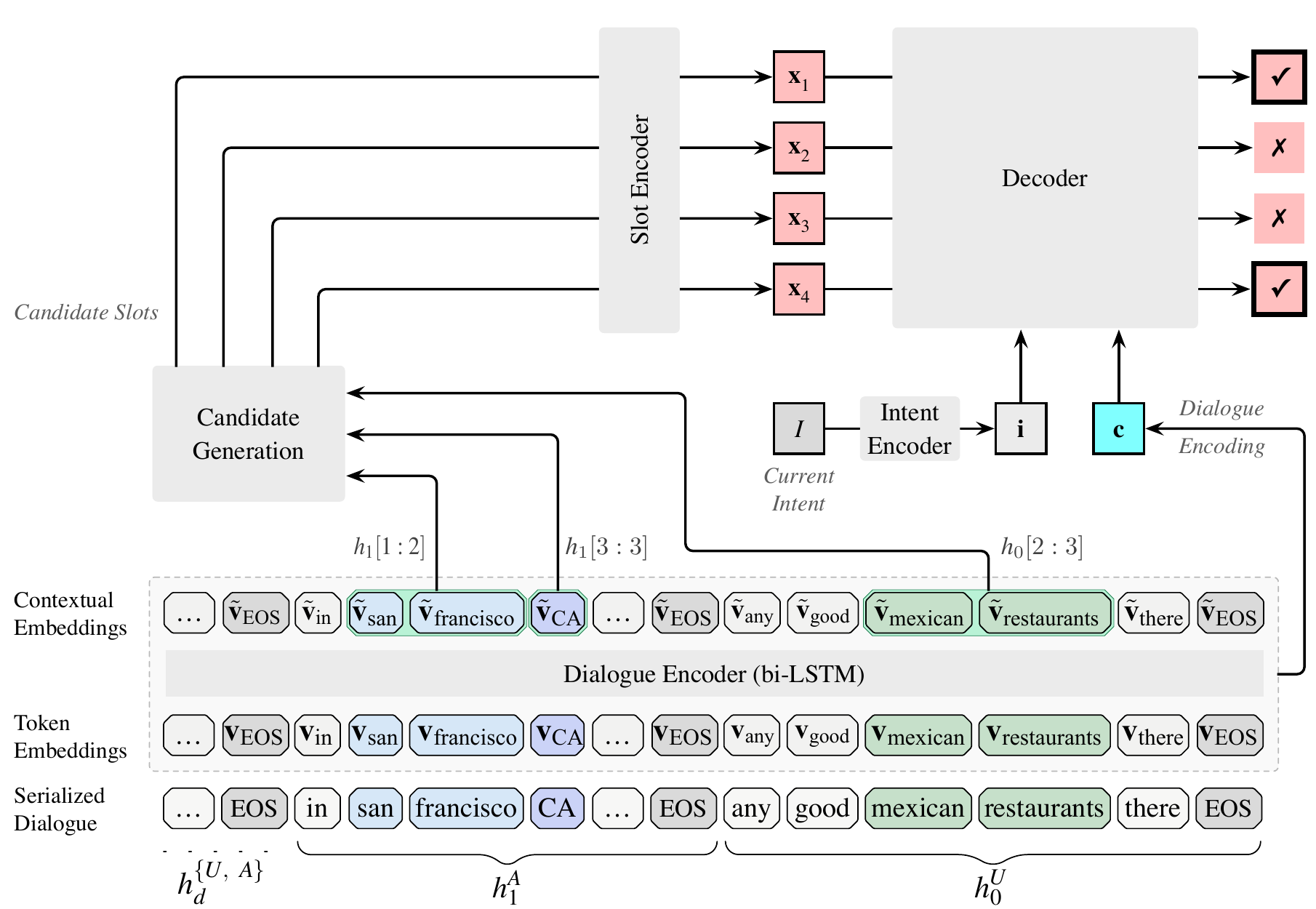}
  \caption{
    General architecture of the proposed contextual carryover model.
    Bi-LSTM is used to encode the utterances in dialogue into a fixed length dialogue representation and also get
    contextual slot value embeddings. Slot encoder uses the slot key, value and distance to create a fixed length slot
    embedding for each of the candidate slots. Given the encoded slots, intent and dialogue context, decoder selects the
    subset of slots that are relevant for the current user request. 
  }
  \label{fig:arch}
\end{figure*}

\paragraph{Candidate Generation}
We follow the approach in~\citet{naik2018contextual}, where, given a dialogue $H$, we construct a candidate set of slots $X$ from the context by leveraging the slot key embeddings to find the nearest slot keys that are associated with the current turn.

\paragraph{Slot Encoder}
A model, given a candidate slot (a slot key, a span in the history and a distance), results in a fixed-length vector
representation of a slot: $\bfx = F_{\rm S}(x, H) \in \mathbb{R}^{D_{\rm S}} $, where $x$ is the slot, $H$ is the full
history.

\paragraph{Dialogue Encoder}
We serialize the utterances in the dialogue and use BiLSTM to encode the context as a fixed-length vector
$ \bfc =~\text{BiLSTM}(H) \in \mathbb{R}^{D_{\rm C}}$.

\paragraph{Intent Encoder}
The intent $I$ of the most recent utterance determined by an NLU module is also encoded as a fixed-length vector
$\bfi \in \mathbb{R}^{D_{\rm I}} $ by averaging the tokens in the intent. We average the word embeddings of the tokens associated with the intent to get the intent embedding.

\paragraph{Decoder}
Given the encoded vector representations $\{ \mathbf{x}_1, \cdots, \mathbf{x}_n \}$ of the slots, the context
vector~$\mathbf{c}$, the intent vector $\mathbf{i}$, produce a subset of the slot ids:
\begin{equation}
  F_{\rm D}(\bfx_{1:n}, \bfc, \bfi) \subseteq \{1, \cdots, N\}
\end{equation}

The overall architecture of the model is shown in \autoref{fig:arch}. We elaborate on the specific designs of these
components under this general architecture.

\subsection{Slot Encoder Variants}
\label{ssec:encoders}
In this section, we describe the different encoding methods that we use to encode slots.

We average the word embeddings of the tokens in the slot key as the \emph{slot key encoding}:
\begin{equation}
\label{eq:slot-key-encoding}
  \bfx_{\rm key}=\frac{1}{K}\sum_{i=1}^K {\bfv(k_i)} \ .
\end{equation}
where $\bfv(w)$ is the embedding vector of token $w$.

For the slot value (the tokens $h_d[l:r]$), we propose following encoding approaches.
\paragraph{CTX\textsubscript{avg}}
The first is to average the token embeddings of the tokens in the slot value:
\begin{equation}
\label{eq:slot-val-avg-encoding}
  \bfx_{\rm val}=\frac{1}{r-l+1}\sum_{i=l}^{r}{\bfv(h_d[i])} \ ;
\end{equation}

\paragraph{CTX\textsubscript{LSTM}}
To get improved contextualized representation of the slot value in dialogue, we also use neural network models to encode
slots. We experimented with bidirectional LSTM \cite{hochreiter1997long} model for slot encoding. LSTMs are equipped
with feedback loops in their recurrent layer, which helps store contextual information over a long history. We encode
all dialogue utterances with BiLSTM to obtain contextualized vector representations $\tilde \bfv(w)$ for each token $w$,
then average the output hidden states of the tokens in the span $[l:r]$ to get the slot value encoding.

\begin{equation}
\label{eq:slot-val-ctx-encoding}
  \bfx_{\rm val}=\frac{1}{r-l+1}\sum_{i=l}^{r}{\tilde\bfv(h_d[i])} ;
\end{equation}

Additionally, \emph{distance} may contain important signals. This integer, being odd or even, provides information on
whether this utterance is uttered by a user or the system. The smaller it is, the closer a slot is to the current
utterance, hence implicitly more probable to be carried over. Building on these intuitions, we encode the distance as a
small vector ($\bfx_{\rm dist}$, 4 dimensions) and append it to the overall slot encoding:
\begin{equation}
\label{eq:slot-encoding}
  \bfx = \left[ \bfx_{\rm key} ~;~ \bfx_{\rm val} ~;~ \bfx_{\rm dist} \right] .
\end{equation}

\subsection{Decoder Variants}
\paragraph{Pointer network decoder}
\label{ssec:pointer_network_decoder}
\begin{figure}[ht]
  \centering
  \includegraphics[width=7cm]{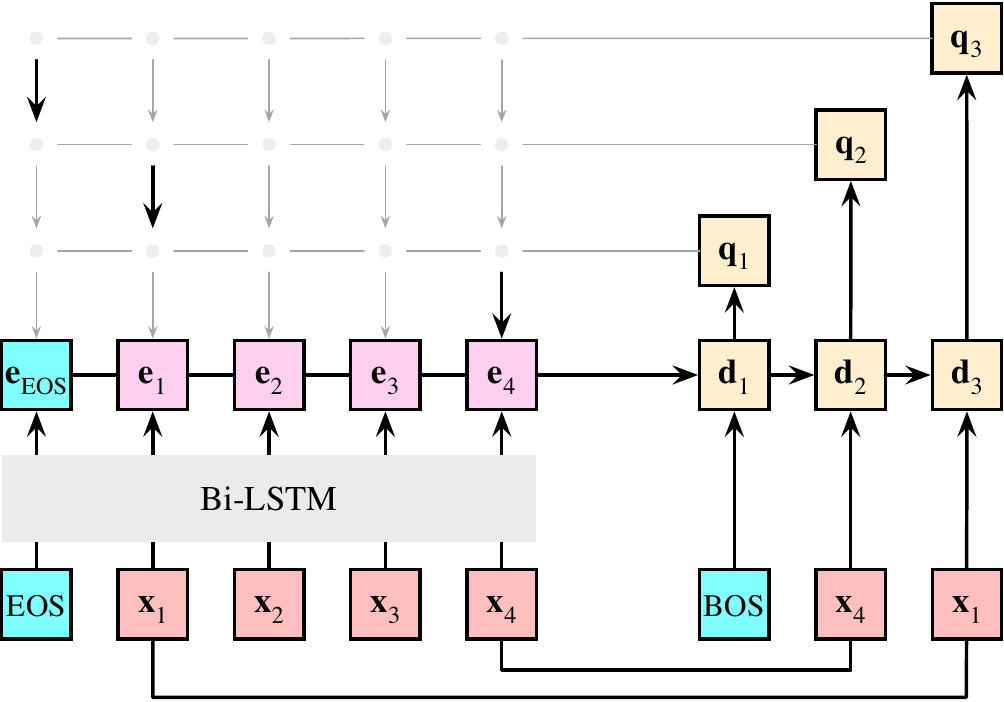}
  \caption{Architecture of the pointer network decoder. In this case, the pointer network selects $\bfx_4$, $\bfx_1$
  successively and stops after selecting $\eos$. }
  \label{fig:ptrnet}
\end{figure}

We adopt the architecture of the pointer network \cite{vinyals2015pointer} as a method to perform joint prediction of the
slots to be carried over. Pointer networks, a variant of Seq2Seq \cite{bahdanau2014neural,sutskever2014sequence,luong2015effective}
model, instead of transducing the input sequence into another output sequence, yields a succession of soft pointers
(attention vectors) to the input sequence, hence producing an ordering of the elements of a variable-length input
sequence.

We use a pointer network to select a \emph{subset} of the slots from the input slot set. The input slot encodings are
ordered as a sequence, then fed into a bidirectional LSTM encoder to yield a sequence of encoded hidden states. We
experiment with different slot orderings as described in section~\ref{sec:experiments}.

\begin{equation}
  \bfe_{0:n} = \mathrm{BiLSTM}([\bfx_\eos ~,~ \bfx_{1:n}])
\end{equation}
Here a special sentinel token \eos ~is appended to the beginning of the input to the pointer network -- when decoding,
once the output pointer points to this \eos ~token, the decoding process stops.

Given the hidden states, $\mathbf{e}_{0:n}$, the decoding process at every time step $i$ is computed and updated as
shown in \autoref{alg:ptrnet}.

\begin{algorithm}[t]
  \caption{Pointer network decoding}
  \label{alg:ptrnet}
  \begin{algorithmic}[1]
    \Procedure{PtrNetDec}{$\bfx_{0:n}, \bfe_{0:n}, \bfd_0, \bfc, \bfi$}
      \State $i \gets 0$
      \State $y_0 \gets \bos$ \Comment{special \bos ~ token}
      \State $ m_{0:n} \gets \boolT $ \Comment{every slot is available}
      \Repeat
        \State $i \gets i + 1$
        \State $\bfd_{i} \gets \mathrm{LSTM}(\bfd_{i-1}, \bfx_{y_{i-1}})$ \Comment{update state}
        \State $\bfq_i \gets F_{\rm Q}(\bfd_i, \bfc, \bfi)$ \Comment{constructs query}
        \State $a_{ij} \gets F_{\rm A}(\bfq_i, \bfe_j)$ \Comment{attention scores}
        \State $p_{ij} \gets \dfrac{\exp a_{ij}}{\displaystyle\sum_{m_j = \boolT}{\exp a_{ij}}}$ \Comment{soft pointer}
        \State $\displaystyle \hat y_i \gets \argmax_{m_j = \boolT} p_{ij} $ \Comment{predicted output}
        \If{at inference time}
        \State $y_i \gets \hat y_i$ \Comment{no gold output}
        \EndIf
        \State $m_{y_i} \gets \boolF$ \Comment{update mask}
       \Until{$ y_i = 0 $} \Comment{index of \eos ~is 0} \\
      \Return $\hat y_{1:i-1}$ \Comment{return all generated $\hat y$'s}
    \EndProcedure
  \end{algorithmic}
\end{algorithm}

Contrary to normal attention-based models which directly uses the decoder state ($\bfd_i$) as the query, we incorporate
the context vector ($\bfc$) and the intent vector ($\bfi$) into the attention query. The query vector is a concatenation
of the three components:
\begin{equation}
\label{eq:ptrnet-query}
  \bfq_i = F_{\rm Q}(\bfd_i, \bfc, \bfi) = \left[~ \bfd_i ~;~ \bfc ~;~ \bfi ~\right].
\end{equation}

We use the general Luong attention \cite{luong2015effective} scoring function (bilinear form):
\begin{equation}
\label{eq:ptrnet-attn}
  a_{ij} = F_{\rm A}(\bfq_i, \bfe_j) = \bfq_i^{\rm T}\bfW\bfe_j \ .
\end{equation}

As a subset output is desired, the output $\hat y_i$ should be distinct at each step $i$. To this end, we utilize a
\emph{dynamic mask} in the decoding process: for every input slot encoding $\bfx_j$ a Boolean mask variable $m_j$ is
set to \boolT. Once a specific slot is generated, it is crossed out -- its corresponding mask is set to \boolF, and
further pointers will never attend to this slot again. Hence distinctness of the output sequence is ensured.

\paragraph{Self-attention decoder}
\label{ssec:transformer_decoder}
The pointer network as introduced previously yields a succession of pointers that select slots based on attention
scores, which allows the model to look back and forth over entire slot sequence for slot dependency modeling. Similar to
the pointer network, the self-attention mechanism is also capable of modeling
relationships between all slots in the dialogue, regardless of their respective positions. To compute the representation of
any given slot, the self-attention model compares it to every other slot in the dialogue. The result of these
comparisons is attention scores which determine how much each of the other slots should contribute to the representation
of the given slot. In this section, we also propose to use the self-attention mechanism with the neural transformer
networks~\cite{vaswani2017attention} to model slot interdependencies for the task.

One major component in the transformer is the multi-head self-attention unit. Rather than only computing the attention
once, the multi-head mechanism runs through the scaled dot-product attention multiple times and allows the model to
jointly attend to information from different perspectives at different positions, which is empirically shown to be more
powerful than a single attention head~\cite{vaswani2017attention}. In our configurations, we increase the number of
heads $Z$, as described in section~\ref{sec:experiments}. The independent attention head $g$ outputs are simply
concatenated and linearly transformed into the expected output.

Given the input slot encodings $\bfx_{1:n}$, we compute the self-attention as follows:
\begin{align}
  \bfq_i^{z} &= \bfW_{\rm Q}^{z}  F_{\rm Q}(\bfx_i) \label{eq:selfattn-query} \\
  \bfk_i^{z} &= \bfW_{\rm K}^{z}  \bfx_i \\
  a_{ij}^{z} &= F_{\rm A}(\bfq_{i}^{z}, \bfk_{j}^{z}) \label{eq:selfattn-attn} \\
  p_{ij}^{z} &= \dfrac{\exp a_{ij}^{z}}{\displaystyle\sum_j \exp a_{ij}^{z}} \\
  {\bfo}_i^{z} &= \sum_j p_{ij}^{z} \bfk_{j}^{z} \\
  \bfxt_i &= \bfW_{\rm O} \left[ {\bfo}_i^0 ~;~ \cdots ~;~ {\bfo}_i^{Z-1} \right] + \bfb_{\rm O}
\end{align}

\noindent where the superscript $0 \le z < Z$ is the head number. We model the query construction,
\autoref{eq:selfattn-query}, and the attention score, \autoref{eq:selfattn-attn}, in the same way as their counterparts
(\autoref{eq:ptrnet-query} and \autoref{eq:ptrnet-attn}) in the previous pointer network model. The self-attended
representation of slot $i$, $\bfxt_i$, is a representation of slot $i$ with the relations to all other slots taken into
account.

We derive the final decision over whether to carry over a slot as a 2-layer feedforward neural network atop the features
$\bfx_i$, $\bfxt_i$, context vector $(\bfc$) and the intent vector ($\bfi$):
\begin{equation*}
  y_i = \nonumber \\
  \sigma(\bfW_2 \cdot \mathrm{ReLU}(\bfW_1 [~ \bfx_i ~;~ \bfxt_i ~;~ \bfc ~;~ \bfi ~] + \bfb_1) + \bfb_2) \ .
\end{equation*}
This creates a highway network connection \cite{srivastava2015highway} that connects the input and the
self-attention transformed encodings.

\biblio


\section{Experiments}
\label{sec:experiments}

\subsection{Datasets}

\begin{table}[t!]
 \renewcommand{\arraystretch}{0.9}
  \begin{center}
  \resizebox{0.48\textwidth}{!}{
       \begin{tabular}{llcccc}
           \toprule
           \multicolumn{2}{c}{\multirow{2}{*}{\bf Split}}  & \multicolumn{4}{c}{\bf Slot distance} \\
           \cmidrule{3-6}
           \multicolumn{2}{l}{}               & \bf 0  & \bf 1  & \bf 2 & \bf $\ge$3 \\
           \midrule
           \multirow{2}{*}{Train}   & Positive  & 183K  & 48K   & 6.7K  & 591   \\
                                    & Total     & 183K  & 327K  & 111K  & 108K  \\
           \midrule
           \multirow{2}{*}{Dev}     & Positive  & 22K   & 6.0K  & 785   & 66    \\
                                    & Total     & 22K   & 40K   & 13K   & 13K   \\
           \midrule
           \multirow{2}{*}{Test}    & Positive  & 23K   & 6.1K  & 807   & 85    \\
                                    & Total     & 23K   & 41K   & 13K   & 14K   \\
           \bottomrule
       \end{tabular}
   }
  \end{center}
  \caption{
    {\bf Internal Dataset} breakdown showing the number of carryover candidate slots at different distances.
    `Total' shows the total number of candidate slots and `Positive' shows the number of candidate slots that are
    relevant for the current turn.
 }
   \label{tbl:internal_candidate_data}
\end{table}

\begin{table}[t!]
 \renewcommand{\arraystretch}{0.9}
  \begin{center}
  \resizebox{0.48\textwidth}{!}{
       \begin{tabular}{llcccc}
           \toprule
           \multicolumn{2}{c}{\multirow{2}{*}{\bf Split}}  & \multicolumn{4}{c}{\bf Slot distance} \\
           \cmidrule{3-6}
           \multicolumn{2}{l}{}               & \bf 0  & \bf 2  & \bf 4 & \bf $\ge$6 \\
           \midrule
           \multirow{2}{*}{Train}   & Positive  & 4.6K  & 3.8K  & 3.7K  & 9.6K   \\
                                    & Total     & 5.2K  & 4.9K  & 4.7K  & 14.5K  \\
           \midrule
           \multirow{2}{*}{Dev}     & Positive  & 1.4K  & 1.2K  & 1.1K  & 3.0K   \\
                                    & Total     & 1.7K  & 1.6K  & 1.5K  & 5.0K   \\
           \midrule
           \multirow{2}{*}{Test}    & Positive  & 4.1K   & 3.2K  & 3.0K  & 9.4K  \\
                                    & Total     & 4.8K   & 4.2K  & 3.9K  & 15.2K \\
           \bottomrule
       \end{tabular}
   }
  \end{center}
  \caption{
    {\bf DSTC2 Dataset} breakdown showing the number of carryover candidate slots at different distances.
    `Total' shows the total number of candidate slots and `Positive' shows the number of candidate slots that represent
    the user goal at the current turn.
  }
   \label{tbl:dstc_candidate_data}
\end{table}

We evaluate our approaches on both internal and external datasets. The internal dataset contains dialogues collected
specifically for reference resolution, while the external dataset was collected for dialogue state tracking.

\paragraph{Internal}
This dataset is made up of a subset of user-initiated dialogue data collected from a commercial voice-based digital
assistant. This dataset has $156$K dialogues from $7$ domains -- Music, Q\&A, Video, Weather, Local Businesses and Home
Automation. Each domain has its own schema. There are ${\sim}13$ distinct slot keys per domain and only $20\%$ of these
keys are reused in more than one domain. To handle dialogue data belonging to a diverse schema, slots in dialogue are
converted into candidate slots in the schema associated with the current domain. We follow the same slot candidate
generation recipe by leveraging slot key embedding similarities as in~\citet{naik2018contextual}. These candidates are
then presented to the models for selecting a subset of relevant candidate slots.
Statistics for the candidate slots in the train, development, and test sets broken down by slot distances are shown in
Table~\ref{tbl:internal_candidate_data}.

\paragraph{DSTC2}
The DSTC2 dataset~\citep{henderson2014word} contains system-initiated dialogues between human and dialogue systems in
restaurant booking domain. We use top ASR hypothesis as the user utterance and use all the slots from n-best SLU with
score $>0.1$ as candidate slots. These candidates are then presented to the models for selecting a subset of candidate
slots which represent the user goal. Statistics for the candidate slots in the train, development, and test sets broken
down by slot distances are shown in Table~\ref{tbl:dstc_candidate_data}. Since only the user mentioned slots contribute
to the user-goal, there are no candidates with odd-numbered slot distances. 

\begin{table*}[t]
 \renewcommand{\arraystretch}{1.1}
   \begin{center}
   \resizebox{0.9\textwidth}{!}{
       \begin{tabular}{@{}  L{4.6cm} L{1.8cm} L{2.4cm} M{1.0cm} M{1.0cm} M{1.0cm} | M{1.0cm} @{}}
           \toprule
           \multirow{2}{*}{Decoder} & \multirow{2}{*}{\parbox{1.8cm}{Slot \\Encoder}} &  \multirow{2}{*}{\parbox{2.4cm}{Slot \\Ordering}} & \multicolumn{4}{c}{Slot distance} \\
           \cmidrule{4-7}
           & &                                                                   & \bf 1               & \bf 2               & \bf $\geq$3         & \bf $\geq$1        \\
           \midrule
           \multicolumn{3}{@{} l}{Baseline \citep{naik2018contextual}}           & \bf 0.8818          & 0.6551              & 0.0000              & \underline{0.8506} \\
           \hline
           \multirow{4}{*}{\makecell[l]{Pointer Network \\Decoder}}
              & \parbox{2cm}{CTX\textsubscript{LSTM}} & \emph{no order}          & 0.8155              & 0.5571              & 0.1290              & 0.7817             \\
              & \parbox{2cm}{CTX\textsubscript{LSTM}} & \emph{turn-only order}   & 0.8466              & 0.6154              & \bf 0.4095          & 0.8157             \\
              & \parbox{2cm}{CTX\textsubscript{avg}} & \emph{temporal order}     & 0.7565              & 0.4716              & 0.0225              & 0.7166             \\
              & \parbox{2cm}{CTX\textsubscript{LSTM}} & \emph{temporal order}    & 0.8631              & \underline{0.6623}  & 0.3350              & 0.8318             \\
           \hline
           Transformer \mbox{Decoder}  & CTX\textsubscript{LSTM} &               & \underline{0.8771}  & \bf 0.7035          & \underline{0.3803}  & \bf 0.8533         \\
           \bottomrule
       \end{tabular}
   }
   \end{center}
   \caption{
     Carryover performance (F1) of different models for slots at different distances on Internal dataset. The rightmost
     column contains the aggregate scores for all slots with distance greater than or equal to $1$.
   }
   \label{tbl:internal_eval}
 \end{table*}

\begin{table*}[t]
 \renewcommand{\arraystretch}{1.1}
   \begin{center}
    \resizebox{0.9\textwidth}{!}{
       \begin{tabular}{@{}  L{4.6cm} L{1.8cm} L{2.4cm} M{1.0cm} M{1.0cm} M{1.0cm} M{1.0cm} @{}}
           \toprule
           \multirow{2}{*}{Decoder} & \multirow{2}{*}{\parbox{1.8cm}{Slot \\Encoder}} &  \multirow{2}{*}{\parbox{2.4cm}{Slot \\Ordering}} & \multicolumn{4}{c}{Slot distance} \\
           \cmidrule{4-7}
           & &                                                                   & \bf 0               & \bf 2               & \bf 4               & \bf $\geq$6         \\
           \midrule
           \multicolumn{3}{@{} l}{Baseline \citep{naik2018contextual}}           & 0.9242              & 0.9111              & 0.9134              &  0.8799             \\
           \hline
           \multirow{3}{*}{\makecell[l]{Pointer Network \\ Decoder}}
             & \parbox{2cm}{CTX\textsubscript{LSTM}} & \emph{no order}           & 0.8316              & 0.8199              & 0.8183              & 0.7641              \\
             & \parbox{2cm}{CTX\textsubscript{LSTM}} & \emph{turn-only order}    & 0.9049              & 0.8993              & 0.9145              & 0.8892              \\
             & \parbox{2cm}{CTX\textsubscript{LSTM}} & \emph{temporal order}     & \underline{0.9270}  & \underline{0.9204}  & \bf 0.9290          & \bf 0.9139          \\
           \hline
           Transformer \mbox{Decoder}  & CTX\textsubscript{LSTM} &               & \bf 0.9300          & \bf 0.9269          & \underline{0.9280}  & \underline{0.8949}  \\
           \bottomrule
       \end{tabular}
    }
   \end{center}
   \caption{
     Carryover performance (F1) of different models for slots at different distances on DSTC2 dataset.
   }
   \label{tbl:dstc_eval}
\end{table*}

\subsection{Experimental setup}
For all the models, we initialize the word embeddings using fastText embeddings \citep{lample2017unsupervised}. The
models are trained using mini-batch SGD with Adam optimizer \citep{Kingma2014AdamAM} with a learning rate of $0.001$ to
minimize the negative log-likelihood loss. We set the dropout rate of $0.3$ for our models during training. In our
experiments, we use $300$ dimensions for the LSTM hidden states in the pointer network encoder and decoder. Our
transformer decoder has $1$ layer, $Z=80$ heads, $d_k= d_v= 64$ for the projection size of keys and values in the
attention heads. We do not use positional encoding for the transformer decoder. All pointer network model setups are
trained for $40$ epochs, our transformer models are trained for $200$ epochs. For evaluation on the test set, we pick
the best model based on performance on dev set. We use standard definitions of precision, recall, and F1 by comparing
the reference slots with the model hypothesis slots.

\subsection{Results and discussion}\label{sec:exp_results}
\begin{figure*}[ht!]
     \centering
     \begin{subfigure}[b]{0.24\textwidth}
         \centering
         \includegraphics[width=\textwidth]{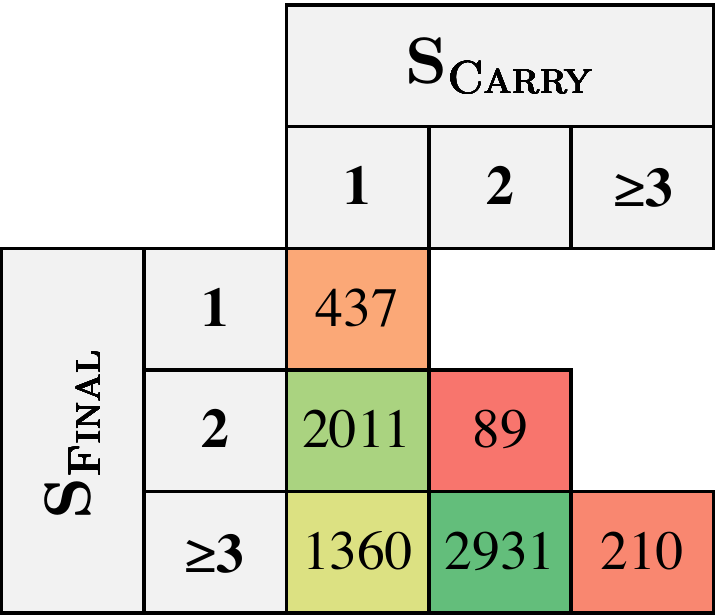}
         \caption{Number of positive instances in the dataset}
         \label{fig:cp_data}
     \end{subfigure}
     \hfill
     \begin{subfigure}[b]{0.24\textwidth}
         \centering
         \includegraphics[width=\textwidth]{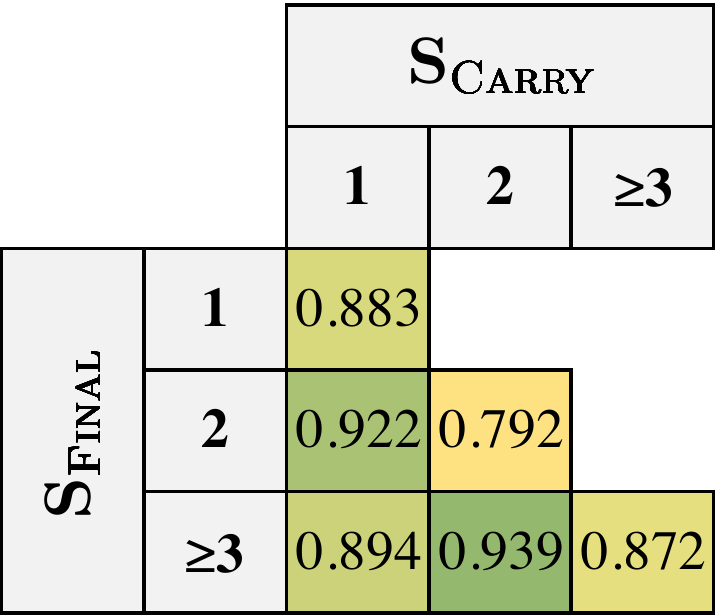}
         \caption{Baseline model performance}
         \label{fig:cp_baseline}
     \end{subfigure}
     \hfill
     \begin{subfigure}[b]{0.24\textwidth}
         \centering
         \includegraphics[width=\textwidth]{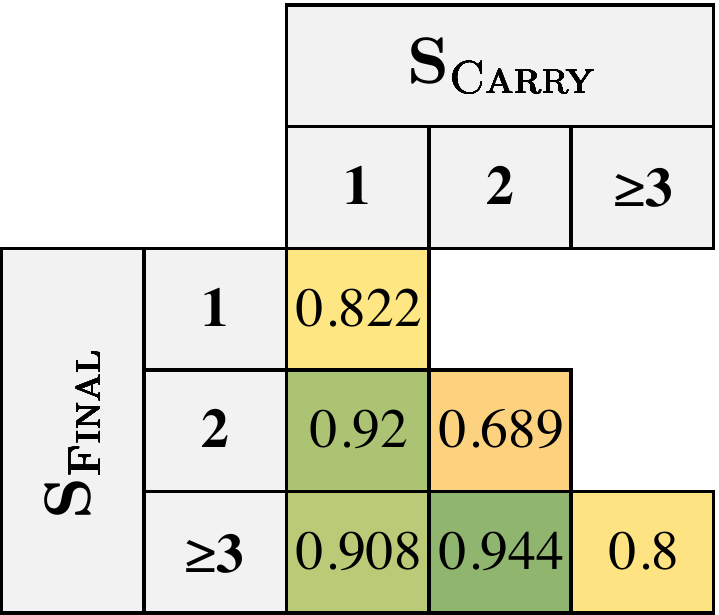}
         \caption{Pointer network performance}
         \label{fig:cp_ptn}
     \end{subfigure}
     \hfill
     \begin{subfigure}[b]{0.24\textwidth}
         \centering
         \includegraphics[width=\textwidth]{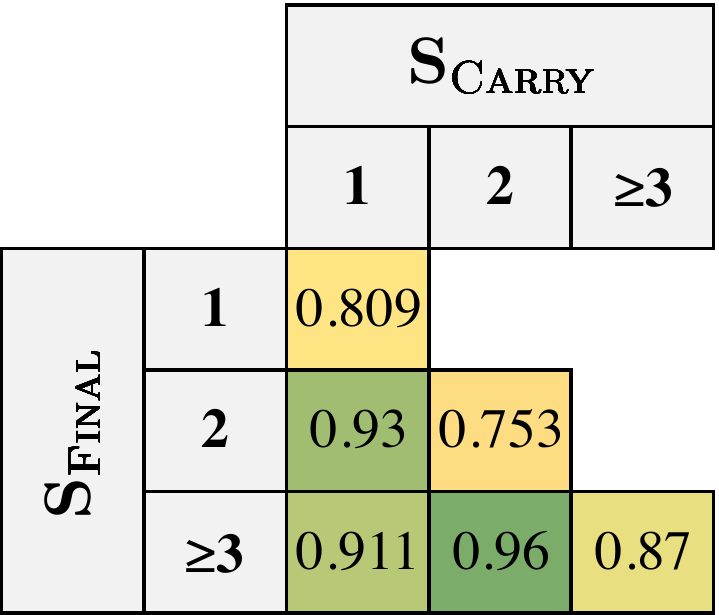}
         \caption{Transformer network performance}
         \label{fig:cp_trn}
     \end{subfigure}
     \caption{
       On internal dataset, plots comparing the performance (F1) of the models across different subsets of candidates
       separated based on the number of final slots after resolution (y-axis) and the number of slots that are carried
       over as part of reference resolution (x-axis)
     }
     \label{fig:analysis}
\end{figure*}

We compare our models against the baseline model -- encoder-decoder with word attention architecture described
by~\citet{naik2018contextual}. Table~\ref{tbl:internal_eval} shows the performance of the models for slots at different
distances on Internal dataset.

\paragraph{Impact of slot ordering}
Using pointer network model, we experiment with the following slot orderings to measure the impact of the order on
carryover performance.
\emph{no order} -- slots are ordered completely randomly.
\emph{turn-only order} -- slots are ordered based on their slot distance, but the slots with the same distance (i.e.,
  candidates generated from the same contextual turn) are ordered randomly.
\emph{temporal order} -- slots are ordered based on the order in which they occur in the dialogue.

Partial ordering slots across turns  i.e., \emph{turn-only order} significantly improves the carryover performance as
compared to using \emph{no order}. Further, enforcing within distance order using \emph{temporal order} improves the
overall performance slightly, but we see drop in F1 by 7 points for slots at distance $\geq$3. indicating that a strict ordering might hurt model accuracy.

\paragraph{Impact of slot encoding}
Here, we compare slot value representations obtained by averaging pre-trained embeddings (CTX\textsubscript{avg}) with contextualized slot value representation obtained from BiLSTM over complete dialogue(CTX\textsubscript{LSTM}). The results in Table~\ref{tbl:internal_eval}, show that contextualized slot value representation substantially improves model performance compared to the non-contextual  representation. This is aligned with the observations on other tasks using contextual word vectors~\citep{Peters2018DeepCW, Ruder2018UniversalLM, Devlin2018BERTPO}.

\paragraph{Impact of decoder}
Compared to the baseline model, both the pointer network model and the transformer model are able to carry over longer dialogue context due to being able to model the slot interdependence. With the transformer network, we completely forgo ordering information. Though the slot embedding includes distance
feature $\bfx_{\rm dist}$, the actual order in which the slots are arranged does not matter. We see improvement in
carryover performance for slots at all distances. While the pointer network seems to deal with longer context better, the transformer architecture still gives us the best overall performance.

For completeness,  Table~\ref{tbl:dstc_eval} shows the performance on DSTC2 public dataset, where similar conclusions hold.

\subsection{Error Analysis}
To gain deeper insight into the ability of the models to learn and utilize slot co-occurrence patterns, we measure the
models' performance on buckets obtained by slicing the data using S\textsubscript{\sc Final} -- total number of slots
after resolution (i.e after context carryover) and S\textsubscript{\sc Carry} -- total number of slots carried from
context. For example, in a dialogue, if the current turn utterance has 2 slots, and after reference resolution if we
carry 3 slots from context, the values for S\textsubscript{\sc Final} and S\textsubscript{\sc Carry} would be 5 and 3
respectively. Figure~\ref{fig:analysis} shows the number of instances in each of these buckets and performance of the
baseline model, the best pointer network and transformer models on the internal dataset. We notice that the baseline model
performs better than the proposed models for instances in the table diagonal (S\textsubscript{\sc Final} =
S\textsubscript{\sc Carry}). These are the instances where the current turn has no slots, and all the necessary slots
for the turn have to be carried from historical context. Proposed models perform better in off-diagonal buckets. We
hypothesize that the proposed models use anchor slots (slots in current utterance having slot distance 0 which are
always positive) and learn slot co-occurrence of candidate slots from context with these anchor slots to improve
resolution (i.e., carryover) from longer distances.

\biblio


\section{Related Work}
\label{sec:related_work}
\begin{figure}[t]
  \centering
  \includegraphics[width=0.48\textwidth]{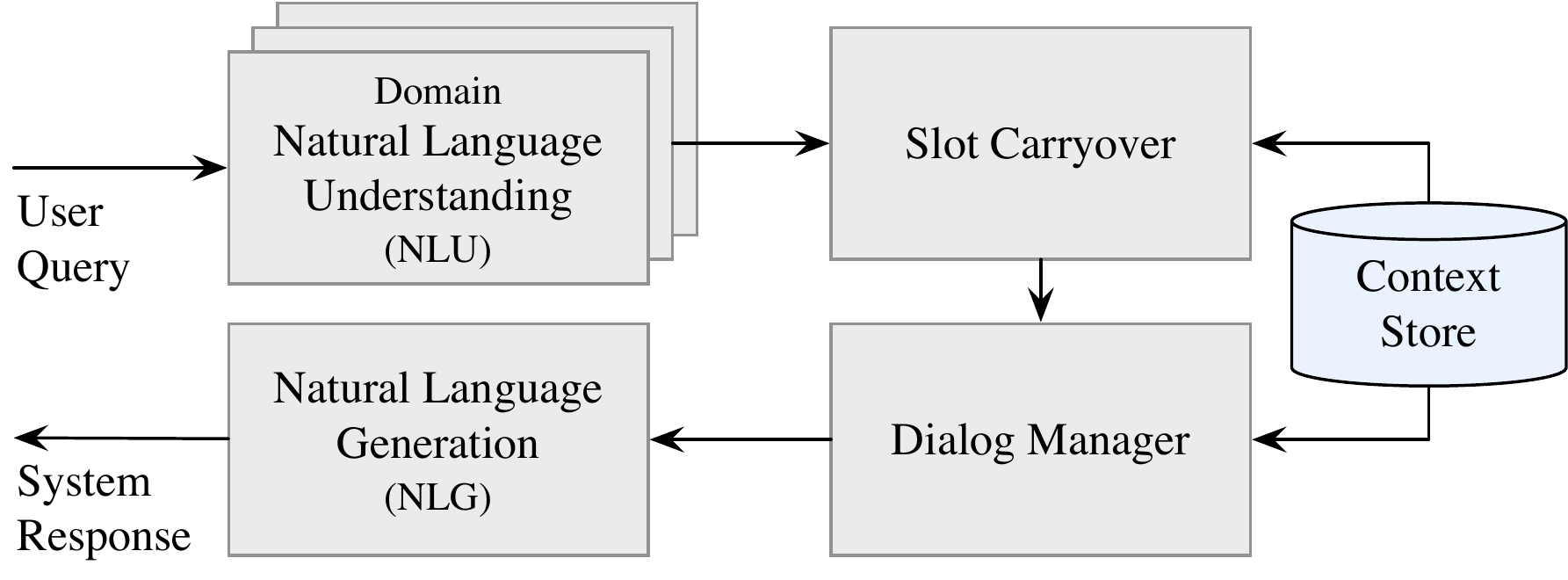}
  \vspace{-0.6cm}
  \caption{
      Spoken dialogue system architecture: the reference resolver/context carryover component is used to resolve
      references in a conversation.
  }
  \label{fig:slu_arch}
\end{figure}

Figure \ref{fig:slu_arch} shows a typical pipelined approach to spoken dialogue~\citep{tur2011spoken}, and where the context carryover system fits into the overall architecture. The context carryover system takes as input, an interpretation output by NLU  -- typically represented as intents and slots \citep{wang2011semantic} -- and outputs another interpretation that contains slots from the dialogue context that are relevant to the current turn. The output from context carryover is then fed to the dialogue manager to take the next action.  Resolving references to slots in the dialogue plays a vital role in tracking conversation states across turns~\citep{celikyilmaz2014resolving}. Previous work, e.g.,
\citet{Bhargava2013EasyCI, Xu2014ContextualDC, bapna2017sequential}, focus on better leveraging dialogue contexts to improve SLU performance. However, in commercial systems like Siri, Google Assistant, and Alexa, the NLU component is a diverse collection of services spanning rules and statistical models. Typical end-to-end approaches \citep{bapna2017sequential} which require back-propagation through the NLU sub-systems are not feasible in this setting.

\paragraph{Dialogue state tracking}
Dialogue state tracking (DST) focuses on tracking conversational states as well. Traditional DST models rely on hand-crafted
semantic delexicalization to achieve generalization \citep{henderson2014word,zilka2015incremental,mrkvsic2015multi}.
\citet{mrkvsic2017neural} utilize representation learning for states rather than using hand-crafted features. These
approaches only operate on fixed ontology and do not generalize well to unknown slot key-value pairs.
\citet{rastogi2017scalable} address this by using sophisticated candidate generation and scoring mechanism while
\citet{xu2018end} use a pointer network to handle unknown slot values. 
\citet{zhong2018global} share global parameters between estimates for each slot to address extraction of rare slot-value
pairs and achieve state-of-the-art on DST. In context carryover, our state tracking does not rely on the definition of user goals and is instead focused on resolving slot references across turns. This approach scales when dealing with multiple spoken language systems, as we do not track the belief states explicitly.

\paragraph{Coreference resolution}
Our problem is closely related to coreference resolution, where mentions in the current utterance are to be detected and linked
to previously mentioned entities. Previous work on coreference resolution have relied on
clustering \cite{bagga1998entity,stoyanov2012easy} or comparing mention pairs
\citep{durrett2013easy,wiseman2015learning,sankepally2018cmr}.
This has two problems.
(1) most traditional methods for coreference resolution follows a pipeline approach, with rich linguistic features,
making the system cumbersome and prone to cascading errors;
(2) Zero pronouns, intent references and other phenomena in spoken dialogue are hard to capture with this approach~\citep{rao2015dialogue}.
These problems are circumvented in our approach for slot carryover.

\biblio

\section{Conclusions}
In this work, we proposed an improvement to the slot carryover task as defined in~\citet{naik2018contextual}. Instead of independent decisions across slots, we proposed two architectures to leverage the slot interdependence -- a pointer network architecture and a self-attention and transformer based architecture. Our experiments show that both proposed models are good at carrying over slots over longer dialogue context. The transformer model with its self attention mechanism gives us the best overall performance. Furthermore, our experiments show that temporal ordering of slots in the dialogue matter, since recent slots are more likely to be referred to by users in a spoken dialogue system. Moreover, contextualized encoding of slots is also important, which follows the trend of contextualized embeddings~\citep{peters2018deep}. 

For future work, we plan to improve these models by encoding the actual dialogue timing information into the contextualized slot embeddings as additional signals. We also plan on exploring the impact of pre-trained representations~\citep{Devlin2018BERTPO} trained specifically over large-scale dialogues as another way to get improved contextualized slot embeddings.

\bibliography{main}
\bibliographystyle{acl_natbib}

\end{document}